\documentclass{article}
\usepackage{ijcai23}

\usepackage{amssymb}
\usepackage{times}
\usepackage{soul}
\usepackage{url}
\usepackage[hidelinks]{hyperref}
\usepackage[utf8]{inputenc}
\usepackage[small]{caption}
\usepackage{graphicx}
\usepackage{amsmath}
\usepackage{amsthm}
\usepackage{booktabs}
\usepackage{algorithm}
\usepackage{algorithmic}
\usepackage[switch]{lineno}
\usepackage{multirow}
\usepackage{ragged2e}
\usepackage{amsmath}
\justifying
\usepackage{float}
\usepackage{xcolor}  %也可以使用{color}宏包，但是颜色库里面颜色较少，使用xcolor会更好
\usepackage{pifont}
\usepackage{CJKutf8}

\title{FORESEE: Multimodal and Multi-view Representation Learning for Robust Prediction of Cancer Survival}

\author{
Liangrui Pan$^1$\and
Yijun Peng$^1$\and
Yan Li$^1$\and
Yiyi Liang$^2$\and
Liwen Xu$^{1}$\and
Qingchun Liang$^{3}$\and
Shaoliang Peng$^1$
\affiliations
$^1$Hunan University
$^2$Shanghai Jiaotong University
$^3$Central South University
\emails
\{panlr, pengyijun,s2310w1062,xuliwen, slpeng\}@hnu.edn.cn,
liangyiyi@renji.com,
503079@csu.edu.cn
}
\begin{document}
\begin{CJK}{UTF8}{gbsn}
%q\linenumbers
\maketitle

\begin{abstract}
%Multimodal medical data plays a crucial role in enhancing the accuracy of survival predictions for cancer patients. Addressing the challenges posed by the specificity and absence of various modalities, this paper introduces a tailored framework, AdaptS, designed to mine multimodal information effectively for robust patient survival prediction. AdaptS incorporates a cross-scale feature cross fusion method to facilitate learning features from pathological images across large, medium, and small fields of view. This enables the acquisition of cross-scale contextual relationships among image features from diverse views. The hybrid attention encoder (HAE) within AdaptS utilizes a contextual attention module to denoise molecular data, capturing both contextual relationship features and local details. Additionally, the channel attention module in HAE enhances the overall representation capability of molecular data. To address intra-modality information loss, we propose asymmetrically masked Triplet masked autoencoder, reconstructing missing information within the modality. Extensive experiments on four cancer cohorts from TCGA demonstrate that our approach consistently outperforms state-of-the-art methods, particularly in scenarios involving modality loss and intra-modality information loss.

Integrating the different data modalities of cancer patients can significantly improve the predictive performance of patient survival. However, most existing methods ignore the simultaneous utilization of rich semantic features at different scales in pathology images. When collecting multimodal data and extracting features, there is a likelihood of encountering intra-modality missing data, introducing noise into the multimodal data. To address these challenges, this paper proposes a new end-to-end framework, FORESEE, for robustly predicting patient survival by mining multimodal information. Specifically, the cross-fusion transformer effectively utilizes features at the cellular level, tissue level, and tumor heterogeneity level to correlate prognosis through a cross-scale feature cross-fusion method. This enhances the ability of pathological image feature representation. Secondly, the hybrid attention encoder (HAE) uses the denoising contextual attention module to obtain the contextual relationship features and local detail features of the molecular data. HAE's channel attention module obtains global features of molecular data. Furthermore, to address the issue of missing information within modalities, we propose an asymmetrically masked triplet masked autoencoder to reconstruct lost information within modalities. Extensive experiments demonstrate the superiority of our method over state-of-the-art methods on four benchmark datasets in both complete and missing settings. %In conclusion, our approach achieves more accurate survival predictions through targeted feature extraction methods for different modality data and handles intra-modal data incompleteness effectively. %We have designed customized feature extraction schemes for each modality, addressing pathology data reflecting morphological features and genomic/transcriptomic data reflecting molecular features. A cross fusion transformer is employed to extract pathology graph structural features at different scales. And a novel hybrid attention encoder (HAE) effectively learns patient molecular features.

\end{abstract}

\section{Introduction}
%Judgment of the survival prognosis of cancer patients plays an important role in clinical decision-making, policy formulation and planning arrangements for patients and their families \cite{liu2021transfer}. Survival prognosis is also a fundamental task in computational pathology, which utilizes tissue whole slide imaging (WSI) for automated risk assessment, patient stratification and triage, and response prediction to treatment. The Cox proportional hazard model is a classic survival prediction model for cancer patients. However, this method fails to account for the role of pathology and genetic material, limiting the effectiveness of survival prediction \cite{christensen1987multivariate}. Secondly, survival prediction analysis significantly differs from cancer staging and grading tasks. It relies on a large amount of data under different factors for analysis, modelling and prediction. Therefore, high-quality acquisition of data from multiple modalities is crucial for development. Patient survival prediction models are important\cite{yao2020whole}. 

\begin{figure}[!t]%[H]
	%\centering
	%\includegraphics[width=0.8\linewidth]{Figure5.pdf}
	\centerline{\includegraphics[width=\columnwidth]{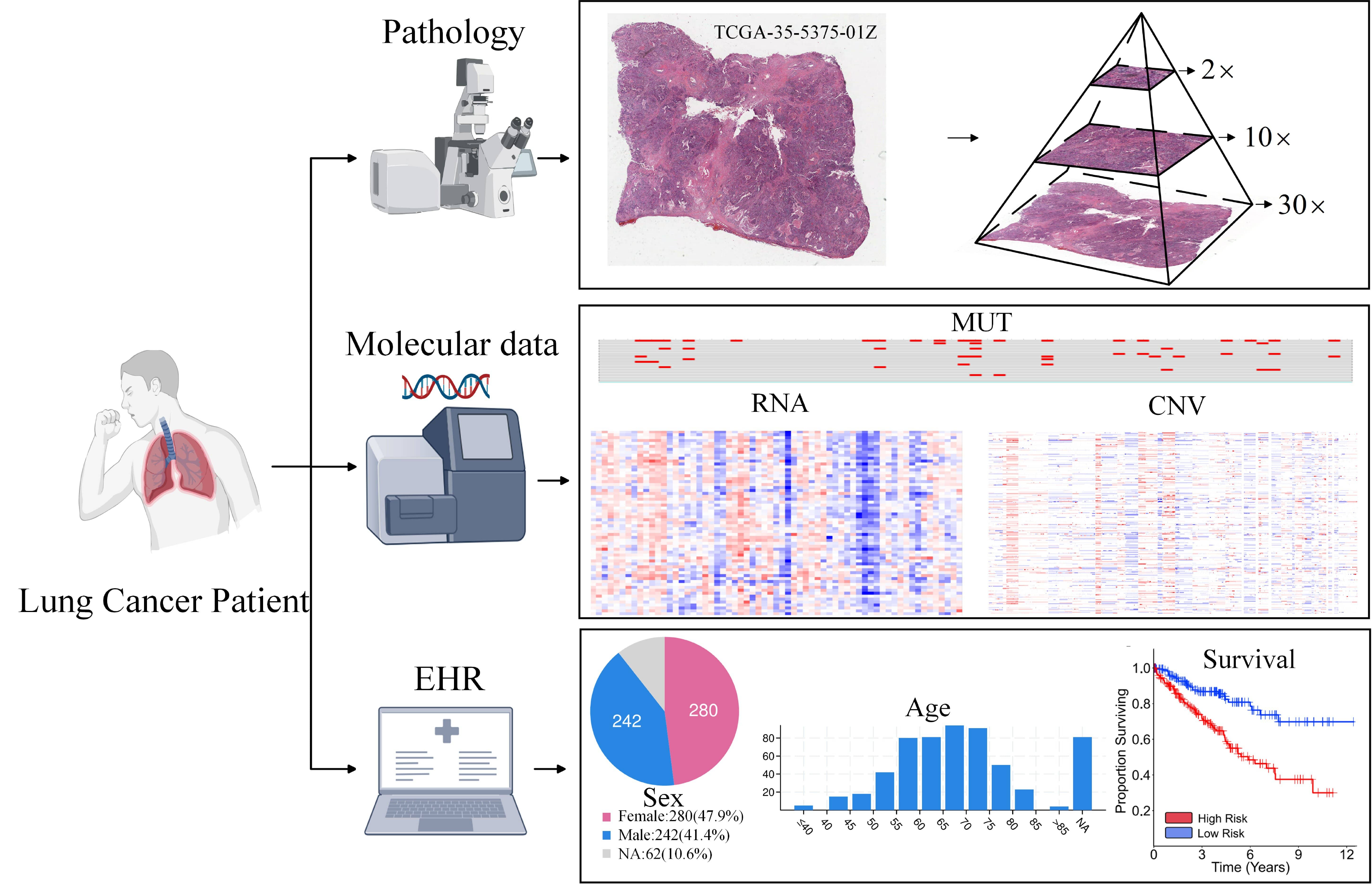}}
	\caption{Obtaining graphical representations of multimodal data (Pathology, Molecular data, and EHR) from lung cancer patients.} %pathology data includes histopathology images at different magnifications. Molecular data primarily consists of RNA, CNV, and MUT. Electronic Health Records mainly encompass age, gender, and survival data.}
	\label{fig:intro}
	%\Description{Pipelines of robust watermarking, fragile watermarking, and our SepMark.}
\end{figure}

With the popularity of third-generation high-throughput sequencing, the cost for doctors to obtain pathological images and molecular data has decreased and the efficiency has been much more accessible \cite{liao2023high}. Illustrated in Figure~\ref{fig:intro}, pathological images can offer semantic information at both cellular and tissue levels at different magnifications. Molecular data includes cancer-related ribonucleic acid (RNA), copy number variation (CNV), and mutation (MUT) data. Due to the rapid advancement of artificial intelligence, deep learning methods utilizing multimodal data are extensively applied in predicting patient survival. Certain technologies, such as AttnMIL \cite{chen2022pan}, GSCNN \cite{mobadersany2018predicting}, Pathomic \cite{chen2020pathomic}, that incorporate feature fusion are proved to be successful in predicting patient survival. However, this method ignores modal incompleteness. It solely enhances feature representation in a unimodal and then employs straightforward feature fusion techniques like vector splicing \cite{mobadersany2018predicting}, modal data aggregation \cite{cheerla2019deep}, Kronecker product \cite{chen2020pathomic} to handle multimodal data. As these methods do not thoroughly discuss the missing modality and the missing information within the modality, the predictive performance of the model will be unreliable in practical applications. % significantly restricted 

The methods based on multimodal data have indeed significantly improved the accuracy of multimodal cancer survival prediction. Unfortunately, these methods neglected the issue of incomplete data in real-life scenarios, arising from patients neglecting certain examinations or being unwilling to conduct them. Hence, the development of a robust model capable of effectively addressing unforeseen missing data in predicting patient survival is essential \cite{wang2022survmaximin}. This alleviates the burden on both physicians and patients. Due to certain constraints, prior methods such as linear interpolation \cite{zhang2017mixup}, matrix factorization \cite{zhang2022robust}, and generative adversarial networks \cite{kaneko2020noise} face challenges when applied to the multimodal survival prediction of cancer patients. In summary, all challenges can be summarized as follows: 1) the heterogeneity and huge dimensional changes of medical multimodal data (e.g., WSI, molecular data) complicate data understanding and processing. Importantly, all methods in survival prediction ignore the simultaneous use of rich semantic features at different scales in WSI. 2) When collecting, organizing and feature extracting multi-modal data, missing modal data and missing data within a modality are prone to occur. 3) Multimodal data obtained through advanced instruments in diverse detection environments inevitably introduce noise \cite{kebschull2015sources}. %Consequently, the substantial disparities in medical data, pose challenges for traditional methods in effectively reconstructing missing modalities or information within modalities.

%Firstly, the substantial dimensional variations in medical data (e.g., WSI, molecular data, electronic health records) complicate data comprehension and processing. Secondly, the heterogeneity of medical data arising from different devices, institutions, and patient groups poses a challenge. Lastly, multimodal data obtained through advanced instruments in diverse detection environments inevitably introduce noise, including polymerase chain reaction (PCR) amplification noise, sequencing errors, and library preparation, etc  \cite{kebschull2015sources}. Consequently, the substantial disparities in medical data, in contrast to natural speech, text, and image data, pose challenges for traditional methods in effectively reconstructing missing modalities or information within modalities. %\iffalse\cite{wang2022survmaximin}\fi

In this paper, we propose a novel end-to-end multimodal framework for robustly predicting patient survival, named FORESEE. To address the heterogeneity of multimodal data and fully mine multimodal features related to prognosis, FORESEE proposes two strategies for extracting information from WSI and molecular data. Firstly, we segment WSI into patches at varying magnification levels and employ a graph structure to characterize multi-scale histopathology data. The cross-fusion transformer (CFT) learns contextual relationships across patches through a cross-scale feature cross-fusion method. This enhances the representation of histopathology images across various scales. Secondly, the novel hybrid attention encoder (HAE) uses a new contextual attention (CTA) module to capture the contextual features of molecular data and extract local feature details. Furthermore, the channel attention (CNA) module in HAE extracts global features of molecular data and improves the overall representation ability of molecular data. The wavelet transform module in HAE resolves the noise present in molecular data. Finally, to tackle the problem of missing intra-modal information, we introduce an asymmetric triplet masked autoencoder (TriMAE) with a non-uniform masks in FORESEE. Specifically, we simulate scenarios with missing data within various modalities. The encoder extracts features exclusively from visible features, while a lightweight decoder reconstructs multimodal data features using latent feature representations and masked tokens. The attention mechanism in the decoder enhances the local details of the reconstructed features. 

Our contributions can be summarized as follows: 1) Based on the spatial topological graph structure of WSI, CFT in FORESEE can effectively cross-fuse features at the cellular level, tissue level, and tumor heterogeneity level. It leverages cross-scale contextual information to enhance the accuracy of survival predictions. 2) The novel HAE utilizes a new contextual attention module and a new channel attention module to learn features of molecular data, reducing the impact of noise in modal data. 3) The asymmetrically masked TriMAE is designed to reconstruct multimodal data features from latent feature representations and masked tokens. 4) End-to-end FORESEE has been extensively and experimentally validated on four cancer datasets of TCGA, achieving the state-of-the-art results and verifying the effectiveness of the above proposed solutions and methods.
%\begin{itemize}
%\item We proposed a CFT that effectively captures local information, global information, and cross-scale contextual information in images at different magnifications. The learning of patch features at smaller magnifications is guided by the learning of patch embeddings at larger magnifications.
%\item The new HAE contains a new contextual attention module and a new channel attention module to learn features of molecular data.
%\item Asymmetrically masked TriMAE is designed to reconstruct multi-modal data features from latent feature representations and masked tokens.
%\item AdaptS has been extensively experimentally verified on four cancer cohorts of TCGA, achieving the best results and verifying the effectiveness of the above proposed solutions and methods.
%\end{itemize}

\section{Related Work}

%\subsection{Multimodal Prediction of Survival}

In recent years, with the continuous emergence of medical big data, clinicians have gradually improved cancer patient survival predictions beyond relying solely on clinical covariates and experience. Computational pathology, nuclear medicine, and third-generation sequencing technologies have made clinicians realize the importance of utilizing artificial intelligence to process multimodal data for supporting cancer survival predictions. In order to further enhance the accuracy of AI-based cancer patient survival prediction, some studies attempt to combine medical imaging data, genomics, and clinical records, proposing research on patient survival prediction based on multimodal data. Currently, most efforts in multimodal survival prediction are primarily focused on feature extraction and data fusion of modality data. Feature extraction from multimodal data commonly uses graph to construct spatial topological structures for image and text data representations. In this context, graph nodes represent feature information, and graph edges signify relationships between node features. For instance, MulGT utilizes a graph transformer architecture to extract low-level representations from graphs \cite{zhao2023mulgt}. HEAT describes WSI as heterogeneous graphs and designs a novel pseudo-label-based semantic-consistent pooling mechanism to obtain graph-level features \cite{chan2023histopathology}. Similarly, MG-Trans employs a patch anchored module (PAM) with multi-head self-attention (MSA) to generate class attention maps for identifying and sampling informative patches \cite{shi2023mg}. Subsequently, multimodal data fusion primarily focuses on late fusion methods, such as Multimodal graph neural network (MGNN) \cite{gao2020mgnn}, GPDBN \cite{wang2021gpdbn}, and Pathomics \cite{chen2020pathomic}. Finally, early fusion methods worth mentioning include the Multimodal co-attention transformer (MCAT) \cite{chen2021multimodal}. These methods effectively improve the accuracy of cancer patient survival predictions.

\section{Methodology}
\begin{figure*}[t]
	\centering
	\includegraphics[width=0.9\linewidth]{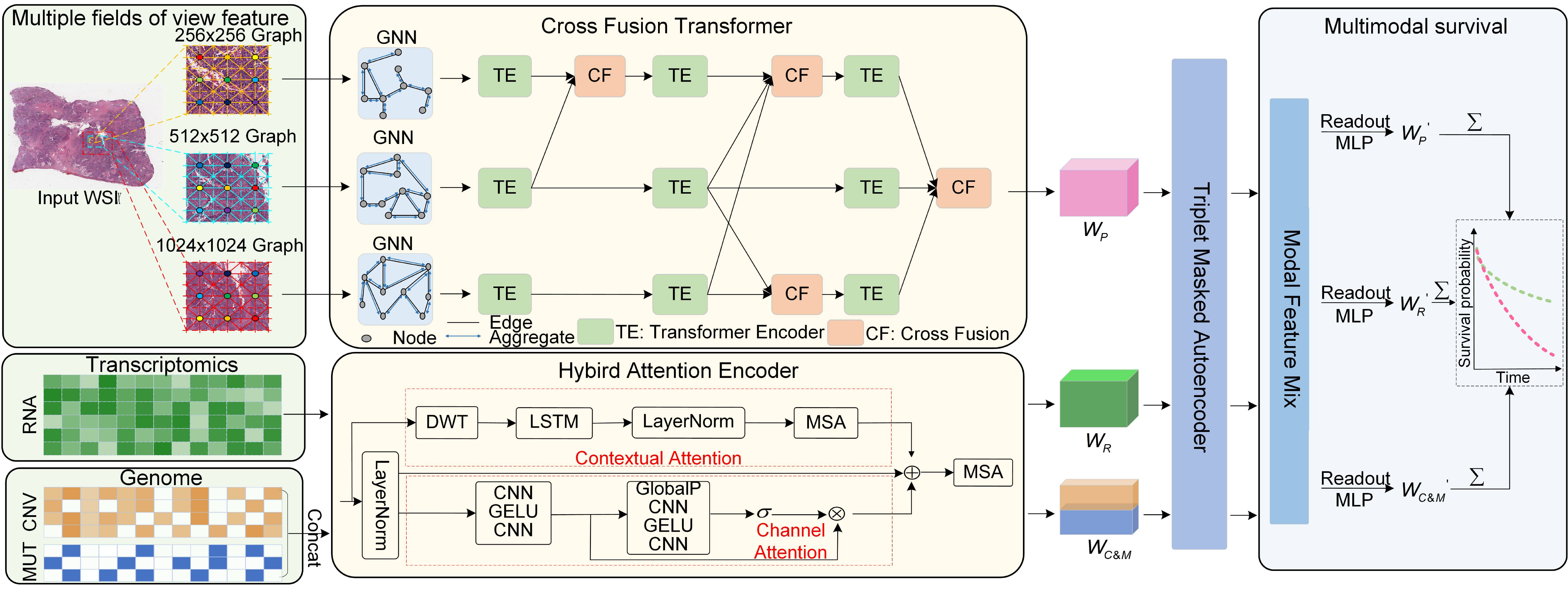}
	\caption{Flowchart depicting multimodal patient survival prediction. The flow chart comprises the multi-view feature extraction , CFT, HAE, TriMAE, and multimodal prediction survival module. Notably, GlobalP denotes the Global Pooling layer, and GELU serves as the activation function.}\label{fig:framwork}
	%\Description{Pipelines of robust watermarking, fragile watermarking, and our SepMark.}
\end{figure*}

\subsection{Overview}
 
Illustrated in Figure~\ref{fig:framwork}, CFT in FORESEE uses a cross-scale feature cross-fusion method to capture different scale feature representations and cross-scale contextual relationships in WSI. To the best of our knowledge, CFT is applied for survival prediction for the first time. HAE extracts local and global information from preprocessed molecular data. TriMAE performs masking operations on different modality data, leveraging feature, mask, and location information to recover potentially lost feature representations. The reconstructed multimodal features are blended and used for final survival prediction.

\subsection{Multimodal Data Preprocessing}
 
To meet the data requirements for predicting patient survival with multimodal data, we selected diagnostic pathology ($P$) data, RNA ($R$), CNV and MUT ($C\&M$) data from TCGA to constitute our multimodal data set $D = \{ P,R,C\&M\}$. %The corresponding observation time and status can be represented as $({t_i},{\beta _i})$.

Illustrated in Figure~\ref{fig:framwork}, WSI predominantly capture cell-level details at low magnification, shifting to tissue-level and tumor heterogeneity level information when observed at high magnification. Therefore, we use a sliding window strategy to segment the WSI into patches of $256\times 256$, $512\times 512$, and $1024\times 1024$, representing small, medium, and large fields of view, respectively. Background is then removed, retaining patches containing image information. To reduce hardware requirements, patch embedding representations need to be extracted before model training. In this work, we use the KimiaNet \cite{riasatian2021fine} to extract patch feature embeddings. Each patch embedding obtained is standardized into a 1024-dimensional vector. Connecting these vectorized feature representations in an 8-neighboring manner yields the graphical representation of pathology images at different scales, denoted as ${G_P} = \{ V,E\}$. where $V$ and $E$ represent the sets of points and edges in the graph, respectively. The post-processed features of WSI are denoted as ${G_{{P_s}}} $ ,${G_{{P_m}}} $ and ${G_{{P_l}}}$. An adjacency matrix $A$ defines the similarity between graph nodes. Secondly, the concat function is employed to merge $C\&M$ data, forming a unified modality. The aforementioned preprocessing involves $P$ data and molecular data, constructing multimodal data consisting of $P$, $R$, and $C\&M$ for each patient. %of size $M \times M$,

\subsection{Cross Fusion Transformer}
%In preprocessing WSI data, the patch input resolution of the transformer is fixed, leading to a limited field of view for the learned features. However,${G_{{P_s}}} $ ,${G_{{P_m}}} $ and ${G_{{P_l}}} $can reflect different local information under various fields of view. A common approach for extracting multi-scale features is to use parallel Transformers to separately extract features for small, medium, and large fields of view. Each transformer contains the same number of encoders. However, this parallel transformers setup has high computational complexity and an extremely large parameter scale, resulting in an explosive increase in hardware costs. To address these issues, we propose the CFT, which efficiently learns cross-scale contextual relationships among encoders of transformers at different fields of view. Compared to parallel transformers, the CFT reduces inference complexity and hardware resource consumption by reducing the model's parameters. In the following, we will introduce the different module functionalities in CFT.

The preprocessed field of view graph ${G_P} = \{ V,E\} $ is a spatial topological diagram of all images in the field of view constructed from a set of nodes $V = \{ {v_1},{v_2}, \ldots ,{v_o}\} $ and the edges $E$ between these nodes. Suppose $({v_m},{v_n}) \in E$ represents an edge from node ${v_m} \in V$ to ${v_n} \in V$, $N({v_m}) = \{ {v_n} \in V|({v_m},{v_n}) \in E\} $ represents the domain of node ${v_m}$, and $\bar N({v_m})$ represents $N({v_m}) \cup \{ {v_m}\} $. The graph ${G_P}$ we designed is undirected, that is, ${v_n} \in N({v_m}) \Leftrightarrow {v_m} \in N({v_n})$. Suppose $N(S) = \{ v \in V|({v_m},{v_n})\} $ represents the domain of node set $S$, and $N(S)$ represents $N(S) \cup S$. Suppose ${x_m} \in {R^{{d_m}}}$  represents the characteristics of node ${v_m}$ with dimension ${d_x}$. ${h_m} \in {R^d}$ is represented as the $d$-dimensional embedding of node ${v_m}$. Let $X = ({x_1},{x_2}, \ldots ,{x_n}) \in {R^{{d_x} \times n}}$ and $H = ({h_1},{h_2}, \ldots ,{h_n}) \in {R^{d \times n}}$. We use ${H_S} = ({h_{{m_k}}})_{k = 1}^{|S|}{R^{d \times |S|}}$ to represent the embedding of the node set $S = ({v_{{m_k}}})_{k = 1}^{|S|}$. $\vec A \in {R^{p \times q}}$ is a vectorized representation of a $p \times q$ matrix $A \in {R^{p \times q}}$, where,${A_{m,n}} \in {A_{m + (n - 1)p}}$. Let ${A_n}$ denote the  $n$-th column of $A$. For positive integers $L$, $[L] = \{ 1,2, \ldots ,L\} $. GNN updates the node’s embedding L by minimizing the objective function. We first use GNN within $L = 2$ layer graph convolution to perform feature extraction on ${G_{{P_s}}} $ ,${G_{{P_m}}} $ and ${G_{{P_l}}}$ to generate the final node embedding $H = {H^L}$, such as:%image features under small, medium and large fields of view to generate the final node embedding $H = {H^L}$, such as: 

\begin{equation}
\begin{array}{l}
	{H^l} = {f_{{W^l}}}({H^{l - 1}};G_{P}),l \in [L],
\end{array}
\end{equation}

When ${H^0} = {G_{P}}$,${f_{{W^l}}}$ is the  $l$-th layer feature transfer function of the learnable parameter ${W^l}$. ${f_{{W^l}}}$ strictly follow the node aggregation and update method. The process is:

\begin{equation}
\begin{array}{l}
	%\begin{array}{l}
		h_m^l = {u_{{W^l}}}(h_m^{l - 1},F_{N({v_m})}^{l - 1},{x_m});\\
		F_{N({v_i})}^{l - 1} = { \oplus _{{W^l}}}(\{ {p_{{W^l}}}(h_n^{l - 1})|{v_n} \in N({v_m})\} ),l \in [L]
	%\end{array}
\end{array}
\end{equation}

Among them, ${p_{{W^l}}}$ can generate a separate feature for each neighbor of ${v_m}$ in the  $l$-th feature transfer iteration. ${ \oplus _{{W^l}}}$ is an aggregation function that maps a set of features to the final feature $F_{N({v_m})}^{l - 1}$. ${u_{{W^l}}}$ is the feature $h_m^{l - 1}$ that is embedded using previous nodes, mapping post feature $F_{N({v_m})}^{l - 1}$ and input feature ${x_m}$ are used to update the optimization function of node embedding. Briefly, we use graph sampling and aggregation in CFT. The channel of each node is mapped to the hidden dimension, which is set to 500. The use of GNN establishes mutual communication between nodes in the feature map ${G_P}$ under each field of view. Then, it saves the extracted features into ${G_P}^{'}$ and passes it to the transformer encoder.

Each transformer learns representative features through a MSA mechanism. The transformer encoder comprises a MSA layer, LayerNorm (LN) layer, and feed-forward network (FFN) layer. Initially, the transformer encoder uses a MSA mechanism to extract abundant semantic information from ${G_P}^{'}$, denoted as ${G_P}^{''}$. To prevent gradient vanishing and enhance model performance, a residual connection is established to transfer information and fuse features ${G_P}^{'}$ and ${G_P}^{''}$. Subsequently, the LN layer is employed to standardize the output, and the FFN executes a non-linear transformation on complex semantic features to capture intricate patterns and features in the input data. To enhance the representation of semantic information and spatial relationships within these modules using CFT, we employ two transformer encoders to learn features for different fields of view.%The process involves:

%\begin{scriptsize}
%\begin{equation}
%\begin{array}{l}
	%\begin{array}{l}
%	{G_P}^{''} = FFN(LN(MSA({G_P}^{'}) + {G_P}^{'}))\\
	%FFN(x) = \max (0,x{W_1} + {b_1}){W_2} + {b_2}\\
%	MSA({G_{{P_q}}}^{'},{G_{{P_k}}}^{'},{G_{{P_v}}}^{'}) = soft\max (\frac{{{G_{{P_q}}}^{'},{G_{{P_k}}}{{^{'}}^T}}}{{\sqrt d }}){G_{{P_v}}}^{'}
	%\end{array}
%\end{array}
%\end{equation}
%\end{scriptsize}

To uncover cross-scale contextual correlations among distinct regions in histopathology images corresponding to different fields of view, we propose to use two types of cross-fusion schemes to blend features from different fields of view. In CFT, the first cross-fusion scheme combine ${G_{{P_s}}}^{''}$ with  ${G_{{P_m}}}^{''}$, ${G_{{P_m}}}^{''}$ with ${G_{{P_l}}}^{''}$, while the second cross-fusion scheme integrates features from large, medium, and small fields of view. Illustrated in Figure~\ref{fig:framwork}, to facilitate the transformer encoder in learning cross-scale context correlations, remapping one-dimensional features ${G_{{P_l}}}^{''} \in {R^{{N^s} \times D}}$ and ${G_{{P_m}}}^{''} \in {R^{{N^s} \times D}}$ is necessary, followed by their integration into a shared feature space denoted as ${G_{{P_{l + m}}}}^{''} \in {R^{({N^s} + {N^s}) \times D}}$. To achieve this objective, the reshaping function $R(.)$ is employed to convert one-dimensional feature maps ${G_{{P_l}}}^{''}$ and ${G_{{P_m}}}^{''}$ into two-dimensional feature maps, followed by splicing operations. Subsequently, two-dimensional convolution is applied for multi-channel attention nonlinear mapping on the spliced features, leading to their fusion. The fused features undergo the upsampling operation of the CNN to maintain consistency between the output and input features. Subsequently, the cross-fusion features are fed into the channel global average pooling and fully connected layers to optimize the output results. Additionally, to avoid potential information loss during intersection, we apply weighting to the sum of the fused features and nonlinear mapping features, ensuring comprehensive integration of semantic information from both large and medium fields of view. The process involves:

{\small
%\begin{scriptsize}
\begin{flalign}
	\label{eq2}
	{F_G} = Fusion(R({G_{{P_l}}}^{''}) \oplus R({G_{{P_m}}}^{''})) \oplus R({G_{{P_l}}}^{''}) \oplus R({G_{{P_m}}}^{''})
\end{flalign}
%\end{scriptsize}
}

$ \oplus $ represents the operation of feature splicing, and $Fusion(.)$ denotes the function of fusing features from both large and small fields of view. ${F_G}$ corresponds to the output of feature fusion. Subsequently, the two-dimensional fused features undergo nonlinear transformation, mapping them to a one-dimensional feature space. Additionally, the features ${G_{{P_m}}}^{''}$  and ${G_{{P_s}}}^{''}$ undergo cross-fusion. The initial type of cross-fusion module is also applicable for the fusion of features ${G_{{P_m}}}^{''}$ and ${G_{{P_s}}}^{''}$. To consolidate features from all fields of view, the second type of cross-fusion aims to intermix ${F_{{G_{{P_l}}}}}$,${F_{{G_{{P_m}}}}}$, and ${F_{{G_{{P_s}}}}}$ processed by the second-layer transformer. The process of the second cross-fusion scheme is similar to the first one, and its process is $Fusion(R({F_{{G_{{P_l}}}}}) \oplus R({F_{{G_{{P_m}}}}}) \oplus R({F_{{G_{{P_s}}}}})) \oplus R({F_{{G_{{P_l}}}}}) \oplus R({F_{{G_{{P_m}}}}}) \oplus R({F_{{G_{{P_s}}}}})$.

%\begin{equation}
%{F_G}^' = Fusion(R({F_{{G_{{P_l}}}}}) \oplus R({F_{{G_{{P_m}}}}}) \oplus R({F_{{G_{{P_s}}}}})) \oplus R({F_{{G_{{P_l}}}}}) \oplus R({F_{{G_{{P_m}}}}}) \oplus {F_{{G_{{P_s}}}}}
%\end{equation}
%The process of the second type of cross fusion is similar to the first type, except that the fused data is added.%Specifically, we simultaneously map three one-dimensional feature maps to a two-dimensional feature space, concatenate them into features, and utilize the multi-channel attributes of the two-dimensional CNN to extract local features from the weighted features. %The fused features undergo the upsampling operation of the CNN to maintain consistency between the output and input features. Subsequently, the cross-fused features are fed into the channel global average pooling and fully connected layers to optimize the output results. %The cross-scale embedding produced by these cross-fusion modules facilitates the graph transformer in learning cross-scale contextual correlations among various fields of view. This aids in extracting the most representative features at different fields of view and magnifications, assisting in patient survival prognosis.

%\begin{figure}[H]
%	\centering
%	\includegraphics[width=0.5\linewidth]{cf.pdf}
%	\caption{Flowchart illustrating feature cross-fusion. \ding{192} represents the fusion of three input features. \ding{193} represents the fusion of two input features.}
%	\label{fig:cf}
	%\Description{Pipelines of robust watermarking, fragile watermarking, and our SepMark.}
%\end{figure}

\subsection{Hybird Attention Encoder}

%RNA sequencing data is primarily transcribed from DNA sequences and possesses robust contextual attributes. CNV arises when the copy number of a gene or genome fragment changes within a chromosomal region, potentially involving gene duplication, deletion, or copy number increase. This phenomenon exhibits pronounced regional attributes. MUT exhibit non-uniform distribution and diversity across distinct regions. Additionally, MUTs in certain genes and gene regions may be interconnected. Thus, we devise a HAE for molecular data processing, encompassing a contextual attention (CTA) module and a channel attention (CNA) module, as illustrated in Fig.~\ref{fig:framwork}.

Given the presence of considerable noise in $P$, $R$, and $C\&M$ data, we employ discrete wavelet transform (DWT) in the CAT module to denoise this data. In DWT, we utilize the Daubechies (dbN) wavelets for preprocessing the input molecular data. Daubechies wavelets, being a class of compactly supported wavelet bases, are suitable for handling signals with mutation characteristics, effectively addressing some gene expression changes or mutated genes in genomic data \cite{adewusi2001wavelet}. Through low-pass filter and high-pass filter, the signal is decomposed into approximation coefficients and detail coefficients at different scales (or frequencies).%, as follows:
%\begin{equation}
%{\small
%\begin{multline}
%\begin{array}{l}
%h(t) = \frac{{1 + \sqrt 3 }}{4}(1 + t){e^{ - \frac{t}{2}}}u(t) - \frac{{1 - \sqrt 3 }}{4}(1 - t){e^{ - \frac{t}{2}}}u( - t)
%\end{array}
%\end{multline}
%\end{equation}
%}

The unit function is denoted by $u(t)$. Soft thresholding is applied to the detail coefficients after the wavelet transform, thereby eliminating noise. The wavelet coefficients after soft thresholding are then subjected to the inverse wavelet transform to reconstruct the denoised signal. For feature extraction from molecular sequencing data, LSTM \cite{huang2015bidirectional} has a significant advantage in handling and capturing long-term dependencies in $R$ and $C\&M$ data. In LSTM, the input data at each time step is transformed into a fixed-length vector, representing the extracted features. Meanwhile, to obtain local attention features of molecular data, we propose using a self-attention mechanism to extract these features.% The process is:%LSTM utilizes gated mechanisms, allowing for the selective retention or forgetting of past information, making it effective in handling long-term dependencies and exhibiting robustness against molecular data noise. Additionally, the data is normalized to standardize input data.

%Meanwhile, to obtain local attention features of molecular data, we propose using a self-attention mechanism to extract these features. In the self-attention mechanism, multiple heads divide the molecular data into segments. Each head maps the input segments into a high-dimensional space and computes attention scores to obtain local attention features. Each head independently obtains local features, and at the end, the locally extracted features from each head are concatenated together. We also employ the Dropout function to prevent the model from over-fitting to the training data. %In the self-attention mechanism, multiple heads divide the molecular data into segments, with each head mapping the segments into a high-dimensional space, generating feature weights $Q,K$ and $V$. We use $Q,K$ to calculate attention scores for obtaining attention weights. Then, these weights are multiplied with  $V$, and the results are summed, yielding the local attention features. 

CNA introduces global information from molecular data to calculate the weight of CNA. The attention mechanism can use global information to weight features and activate more features.  Illustrated in Figure~\ref{fig:framwork}, to reduce the computational cost associated with using a fixed-width convolution directly, we compress the number of channels the two convolutional layers to a constant, denoted as $\beta $. The output channels after the first convolution are compressed to $C\beta $, where $C$ is the channel number of the first convolution. Then, the second convolution expands the features to $C$ channels, followed by the adaptive adjustment of channel features using CNA. Finally, MSA is employed from feature extraction on the merged features. The workflow of the entire mixed attention encoder is as follows: %CNA module consists of several standard one-dimensional convolutional layers with GELU activation functions \cite{hendrycks2016gaussian}, along with another CNA.

{\small
	%\begin{equation}
		\begin{align}
			%\begin{split}
				%\centering
				%{x_{CTA}} = MSA(LN(LSTM(DWT({x_{in}}))));\\
				%{x_{CNA}} = CNA(LN({x_{in}}));\\
				%{x_{out}} = MSA(LN({x_{in}}) + {x_{CTA}} + {x_{CNA}})
				%\begin{array}{l}
				{x_{CAT}} &= MSA(LN(LSTM(DWT(LN({x_{R\|C\& M}})))))\\
				{x_{CNA}} &= \alpha CNA(LN({x_{R\|C\& M}}))\\
				{x_{HAE}} &= MSA(LN({x_{R\|C\& M}}) + {x_{CAT}} + {x_{CNA}})
				%\end{array}
			%\end{split}
		\end{align}
	%\end{equation}
}

%Methods               & $P$ & $R$ & $C,M$  & BLCA   & BRCA  &  LUAD   &  UCEC \\
%\resizebox{\linewidth}{!}{  #此处！表示根据根据宽高比进行自适应缩放

\subsection{Triplet Masked Autoencoder}

To ensure that the model is not adversely affected by missing features during inference, we designed a TriMAE to reconstruct the missing feature information in multimodal data. Inspired by SiamMAE \cite{gupta2023siamese}, which can reconstruct detailed features of masked frames within videos.Through the analysis of multimodal data and features, we found that different modalities have missing features. Therefore, we propose a masking strategy in TriMAE, where each branch masks only one modality feature during the processing of multimodal data. The masking rate is set to be over 80\%, making it challenging for features to reconstruct the masked pixels through surrounding features. This compels the encoder to learn meaningful features.

\begin{figure}[h]
	\centering
	\includegraphics[width=0.95\linewidth]{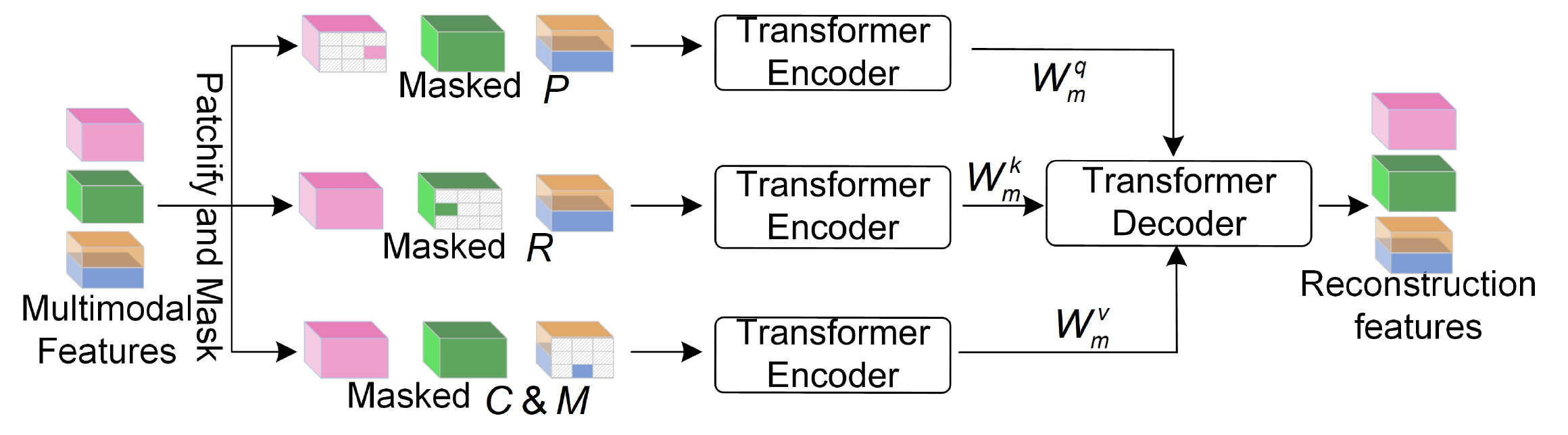}
	\caption{Flowchart for TriMAE to reconstruct missing data in multimodal data.}\label{fig:trimae}
	%\Description{Pipelines of robust watermarking, fragile watermarking, and our SepMark.}
\end{figure}

TriMAE consists of an asymmetric encoder-decoder architecture. The encoder is based on three parallel transformers serving as the backbone network, operating only on unmasked feature patches. This significantly reduces the inference cost of the encoder. The decoder is a lightweight module composed of cross-attention layers and self-attention layers. The feature information obtained by encoders from different branches is focused on each other through a cross-attention layer and then complements missing information by mutually attending to each other through a self-attention layer. The input primarily comprises unmasked feature patches and masked tokens, where the masked features are shared and learned vectors for all missing positions. As illustrated in Figure~\ref{fig:trimae}, TriMAE takes features with masked information from different modalities as input and, through the learning process of the encoder and decoder, reconstructs missing feature details using contextual relationships and semantic information. TriMAE employs the attention mechanism principles of cross-attention layers to perform local attention calculations on reconstructed features from the three branches. After the feature cross calculation, further self-attention feature learning is applied to the reconstructed features to obtain the optimal feature representations for each modality data.%The process is as follows:
%\begin{equation}
%CrossAtten({x_1},{x_2},{x_3}) = soft\max (\frac{{{W^{q}}{W^{k}}^{^{T}}}}{{\sqrt {{d_k}} }}){W^{v}}
%\end{equation}
%The symbols ${x_1},{x_2},{x_3}$ in this context represent features extracted from different branches of TriMAE. ${W^{q}} \in {R^{{d_1} \times {d_k}}}$, ${W^{k}} \in {R^{{d_2} \times {d_k}}}$, and ${W^{v}} \in {R^{{d_3} \times {d_k}}}$ correspondingly denote learned projection matrices. ${d_k}$ is the dimension size of the set of $W_m^{q}$, $W_m^{k}$ and $W_m^{v}$, which is used to scale the attention size. After the feature cross-fusion, further self-attention feature learning is applied to the reconstructed features to obtain the optimal feature representations for each modality data.

\subsection{Training and Inference Strategies}
To establish interaction among multimodal data, we designed a modal feature fusion module to mix information from different modalities and edge information of channels to promote comprehensive interaction between multimodal data. This process is represented as follows:
\begin{equation}
{F} = MLP(LN(MLP(x) + x) + x)
\end{equation}
Finally, the survival output of modality $m$ can be calculated as:
\begin{equation}
{W_m} = MLP({\mathop{\rm Re}\nolimits} adout({F_m}))
\end{equation}
Therefore, the Cox loss of modality $m$ can be defined as:
%\begin{equation}
\begin{small}
\begin{multline}
L_{Cox}^m = \sum\limits_{i = 1}^B {{\delta _i}} \left( { - {O_m}(i) + \log \sum\limits_{j:{t_j} >  = {t_i}} {\exp ({O_m}(j))} } \right)
\end{multline}
\end{small}
%\end{equation}
Where ${O_m}(i)$ and ${O_m}(j)$ represent the survival outputs for the $i$-th and $j$-th patients in modality $m$, respectively. In summary, the loss for the entire model can be defined as:
\begin{equation}
L_{all}^m = \sum\limits_{m \in M} {{\lambda _m}} L_{Cox}^m + {\lambda _0}{L_{TriMAE}}
\end{equation}
Where ${\lambda _0}$ and ${\lambda _m}$ are adjustable parameters. Considering the complexity of multimodal, ${\lambda _0}$ and ${\lambda _m}$ are set to five and one, respectively.

\begin{table*}[h] %*表示居中
	\centering
	\resizebox{0.7\linewidth}{!}{
		\begin{tabular}{l|ccc|cccc}
			\toprule
			%\bottomrule			 
			\textbf{Methods} & \textbf{\boldmath$P$} & \textbf{\boldmath$R$} & \textbf{\boldmath$C\&M$} & \textbf{BLCA} & \textbf{BRCA} & \textbf{LUAD} & \textbf{UCEC}\\			
			\hline
			DeepAttnMISL \cite{yao2020whole}        & \checkmark  &            &            & 0.504±0.042   & 0.524±0.043 & 0.548±0.050      & 0.597±0.059\\
			CLAM-SB \cite{lu2021data}           & \checkmark  &            &            & 0.559±0.034   & 0.573±0.044 &  0.594±0.063    & 0.644±0.061  \\
			CLAM-MB \cite{lu2021data}           & \checkmark  &            &            & 0.565±0.027   & 0.578±0.032 &  0.582±0.072    & 0.609±0.082 \\
			Trans-MIL \cite{shao2021transmil}         & \checkmark  &            &            & 0.575±0.034   & 0.666±0.029 &  0.642±0.047    & 0.655±0.046 \\
			DTFD-MIL \cite{zhang2022dtfd}          & \checkmark  &            &            & 0.546±0.021   & 0.609±0.059 &  0.585±0.066    & 0.656±0.045 \\
			\hline
			Cox \cite{matsuo2019survival}              &             & \checkmark &            & 0.637±0.008         & 0.646±0.021       &  0.621±0.039    & 0.680±0.025  \\		
			SNN-Trans \cite{shao2021transmil}         &             & \checkmark &            & 0.659±0.032   &0.647±0.063  &  0.638±0.022    & 0.656±0.038 \\
			XGBLC \cite{ma2022xgblc}             &             & \checkmark &            & 0.681±0.003   & 0.673±0.006 &  \textbf{0.675±0.004}    & 0.680±0.025 \\
			\hline						
			GSCNN \cite{mobadersany2018predicting}             & \checkmark  & \checkmark &            & 0.596±0.014   & 0.636±0.021 &  0.617±0.012    & 0.693±0.031 \\
			Pathomic \cite{chen2020pathomic}       & \checkmark  & \checkmark &            & 0.586±0.062   & 0.657±0.031       &  0.602±0.014    & 0.676±0.027 \\
			Metric learning \cite{cheerla2019deep}   & \checkmark  & \checkmark &            & 0.584±0.012   & 0.661±0.043 &  0.613±0.026    & 0.690±0.011 \\			
			MultiSurv \cite{vale2021long}       & \checkmark  & \checkmark &            & 0.664±0.012         & 0.656±0.014       &  0.626±0.031    & 0.690±0.015 \\
			MCAT \cite{chen2021multimodal}              & \checkmark  & \checkmark &            & 0.672±0.032   & 0.659±0.031 &  0.659±0.027    & 0.649±0.043 \\
			%Outer product[13]    & \checkmark  & \checkmark &            & \--{}         & \--{}       &  0.602±0.014    & 0.676±0.027  \\
			HGCN \cite{hou2023hybrid}              & \checkmark  & \checkmark &            & 0.675±0.006         & 0.671±0.004      &  0.651±0.008    & 0.682±0.014 \\
			MOTCat \cite{xu2023multimodal}             & \checkmark  & \checkmark &            & 0.683±0.026   & 0.673±0.006 &  0.670±0.038     & 0.675±0.004 \\
			
			AttnMIL \cite{chen2022pan}         & \checkmark  &            &\checkmark  & 0.599±0.048   & 0.609±0.065 & 0.567±0.010      & 0.670±0.011 \\
			
			SNN \cite{chen2022pan}              &             & \checkmark &\checkmark  &0.618±0.022    & 0.624±0.060  &  0.611±0.047    & 0.679±0.040  \\
			\hline
			MMF \cite{chen2022pan}               & \checkmark  & \checkmark & \checkmark & 0.636±0.012   & 0.674±0.023 &  0.600±0.057    & 0.634±0.032 \\
			Porpoise \cite{chen2022pan}         & \checkmark  & \checkmark & \checkmark & 0.636±0.024   & 0.652±0.042 &  0.647±0.031    & 0.675±0.032 \\
			FORESEE                & \checkmark  & \checkmark & \checkmark & \textbf{0.686±0.008}    & \textbf{0.679±0.013} & 0.672±0.013    & \textbf{0.730±0.002} \\
			\bottomrule
	\end{tabular}}
	\caption{Performance of C-Index (mean±std) on four cancer datasets.}
	\label{tab:sota}
\end{table*}

\section{Experiment}
\subsection{Experimental Settings}

%, i.e. the raw data can be represented as WSI-RNA-CNV-MUT

\textbf{Datasets.} Diagnostic WSI and corresponding molecular and clinical data were collected from 2479 patients across BLCA, BRCA, LUAD and UCEC cancer datasets at TCGA through the NIH Genomic Data Commons Data Portal. The availability of CNVs, MUTs and RNA-Seq abundances for each WSI match.  All molecular and clinical data were obtained from cBioPortal’s \cite{cerami2012cbio} quality control files. %This strategy helps optimize the data processing process and improve the efficiency and reliability of analysis.
%\subsection{Evaluation}

\noindent\textbf{Evaluation Metrics.} This experiment employs the C-index and Kaplan-Meier (KM) analysis  method to assess the performance and prediction accuracy of the survival analysis model \cite{xu2023multimodal}. %The C-index value ranges from 0.5 to 1, where 0.5 indicates performance equivalent to random guessing, and 1 indicates perfect prediction. Each vertical step on the KM survival curve signifies the occurrence of an event, with the downward trend reflecting the cumulative event incidence over time. %This article will comprehensively evaluate the model through curve comparison, median survival time, and Log-rank test. Curve comparison assesses the model's ability to differentiate between high- and low-risk groups; median survival time involves categorizing patients into high-risk and low-risk groups based on median scores; a lower P value in the Log-rank test indicates superior model performance.

%\subsection{Implementation Detail}
\noindent\textbf{Implementation.} All experiments are based on NVIDIA GeForce RTX 4090. For the cancer datasets, we perform 5-fold cross-validation and report the cross-validated concordance index (C-Index) and its standard deviation (std). All models are developed using Python 3.9 based on the PyTorch 1.13.1 platform. After extensive experiments, we set some optimal hyper-parameters for the model. We adopt Adam optimizer with initial learning rate of 5 $\times {10^{ - 3}}$ and weight decay of 1 $\times {10^{ - 5}}$, batch size set 50, epoch set 50, and Dropout set 0.2.

\begin{figure*}[!t]%[t]
	\centering
	\includegraphics[width=0.89\linewidth]{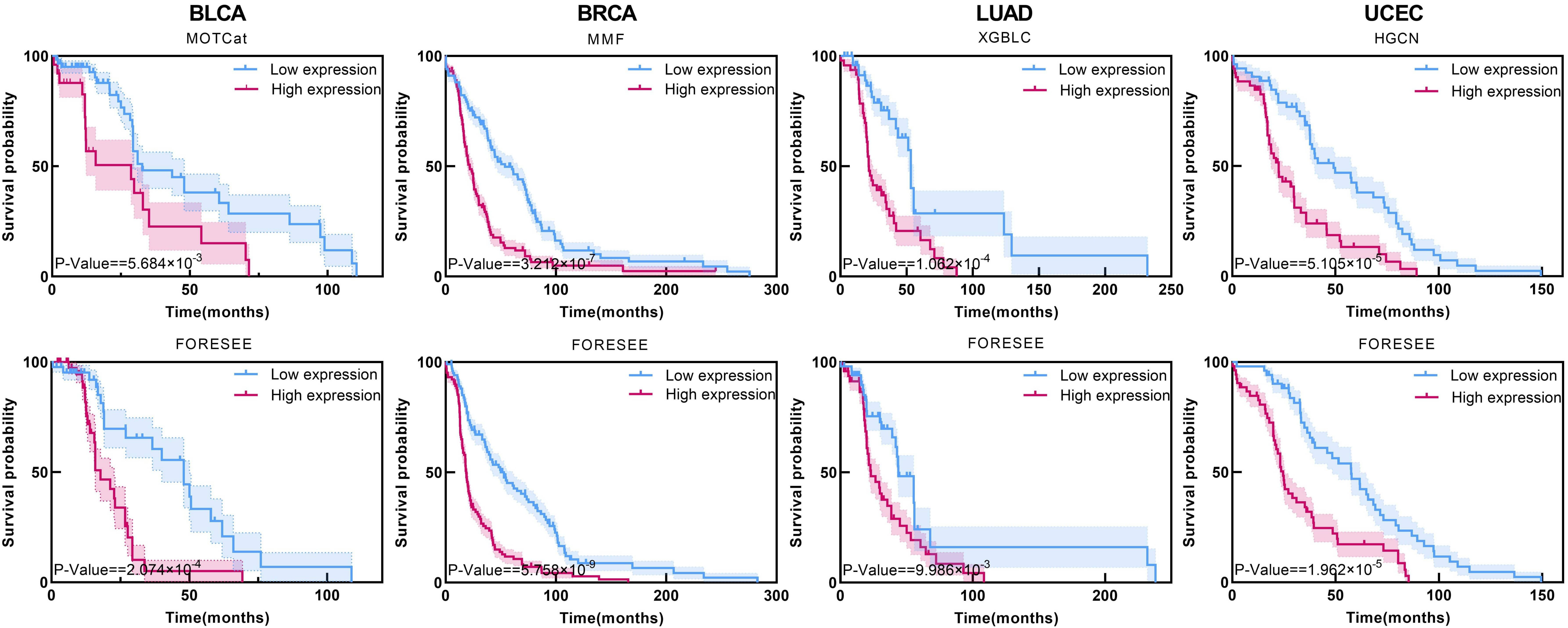}
	\caption{KM analysis compares the performance of state-of-the-art methods with FORESEE tested on four cancer datasets. }\label{fig:survival}
	%\Description{Pipelines of robust watermarking, fragile watermarking, and our SepMark.}
\end{figure*}

\subsection{Comparison With State-of-the-Art Methods}
%\subsection{Quantitative Results}
We compare FORESEE with unimodal and multimodal state-of-the-art (SOTA) methods: for unimodal data, we adopt \textbf{DeepAttnMISL} \cite{yao2020whole}, \textbf{CLAM-SB} \cite{lu2021data}, \textbf{CLAM-MB} \cite{lu2021data}, \textbf{TransMIL} \cite{shao2021transmil}, \textbf{DTFD-MIL} \cite{zhang2022dtfd}, \textbf{Cox} \cite{matsuo2019survival}, \textbf{SNNTrans} \cite{shao2021transmil}, \textbf{XGBLC} \cite{ma2022xgblc}. For multimodal data, we adopted \textbf{GSCNN} \cite{mobadersany2018predicting}, \textbf{Pathomic} \cite{chen2020pathomic}, \textbf{Metric learning} \cite{cheerla2019deep}, \textbf{MultiSurv} \cite{vale2021long}, \textbf{MCAT} \cite{chen2021multimodal}, \textbf{HGCN} \cite{hou2023hybrid}, \textbf{MOTCat} \cite{xu2023multimodal}, \textbf{AttnMIL} \cite{chen2022pan}, \textbf{SNN} \cite{chen2022pan}, \textbf{MMF} \cite{chen2022pan}, \textbf{Porpoise} \cite{chen2022pan}. The results are shown in Table~\ref{tab:sota}. Since different experiments use different cancer data sets, we restore the method in the article and re-experiment and obtain survival prediction results.

\noindent\textbf{Unimodal v.s. Multimodal:} FORESEE achieves the SOTA results in the BLCA, BRCA, and UCEC cancer datasets. This indicates that multimodal approaches can effectively enhance the accuracy of survival prediction, and FORESEE exhibits generalization. Furthermore, results based on the $R$ modality generally outperform those based on the $P$ modality. This is because genes determine phenotypes, and phenotypes are strongly associated with survival. However, XGBLC attains the SOTA result on the LUAD dataset. This suggests that the simple feature concatenation fusion method in FORESEE faces significant challenges on the LUAD dataset.

\noindent\textbf{Multimodal SOTA v.s. FORESEE:} Through a large number of experiments, FORESEE demonstrated an improvement in predictive performance over the SOTA MOTCat by 0.3\%, 0.6\%, and 3.7\% on the BLCA, BRCA, and UCEC datasets, respectively. This indicates that FORESEE enhances the performance of multimodal prediction by incorporating $C\&M$ data, as the expression of $C\&M$ directly signifies the causative factors of cancer. Secondly, the strategy of extracting features from multiple fields of view can effectively assist FORESEE in survival prediction. However, on the UCEC dataset, the XGBLC prediction performance outperformed FORESEE by 0.3\%. Of concern is that the method does not generalize to other cancer datasets. Therefore, mining multimodal data for predicting patient survival will be a trend in future research. %\textcolor{red}{Ablation experiments are in the Appendix}. %However, on the UCEC cohort, the XGBLC prediction performance outperformed AdaptS by 0.3\%. Of concern is that the method does not generalize to other cancer cohorts. \textcolor{red}{Ablation experiments are in the Appendix}.

Next, we categorized each cancer dataset into high and low expression groups based on the median survival risk predicted by our model. Accurate survival predictions should manifest as a significant divergence in the KM curves between these two groups. Figure~\ref{fig:survival} illustrates the superior performance of MOTCat, MMF, XGBLC and HGCN on BLCA, BRCA, LUAD and UCEC datasets, respectively, setting the SOTA benchmarks.The $P$ values of FORESEE in BLCA, BRCA, LUAD and UCEC datasets all surpassed those of the SOTA method, with LUAD patients exhibiting only a slight difference. Nevertheless, the $P$ values of FORESEE across all cancer datasets were consistently below 0.05, affirming the significant value of FORESEE in predicting patient survival.

\subsection{Ablation Study}
\textbf{Effect of \boldmath{${P_s}$}, \boldmath{${P_m}$} and \boldmath{${P_l}$} in \boldmath{${P}$} modality.} To investigate how feature information in large and small fields of view within ${P}$ modality affects the prognosis of cancer patients, we kept modality data $R$ and $C\&M$ constant while systematically altering the fusion strategy for different fields of view.  The results are shown in Table~\ref{tab:com}. Experiments have found that when ${P_s}$ , ${P_m}$ and ${P_l}$ are utilized individually, ${P_s}$  can provide features closely related to prognosis for multimodal data. FORESEE experiments in independent $R$ and $C\&M$  modes respectively found that the prediction performance of using multiple fields of view is better than using a single field of view. FORESEE achieved the best prediction performance on the four cancer datasets with the combination of ${P_s}$ , ${P_m}$. However, using only $C\&M$ and multimodal data composed of different independent fields of view predicts poorer survival, possibly because of the lack of complementary modal information to help multimodal data perform survival analysis. The performance of FORESEE in survival analysis using feature information in all fields of view in $R$ and $C\&M$  modes is significantly better than other cases where field features are missing. In summary, after a large number of experiments, we found that the fusion of features from different fields of view in pathology images is essential for survival prediction. The survival model FORESEE can obtain survival-related features from the cell level, tissue level, and tumor heterogeneity.

\begin{table}[!t]%[h]		
	\renewcommand{\arraystretch}{1}
	\centering
	\resizebox{1\linewidth}{!}{
		\begin{tabular}{c|ccc|cccc}
			%\hline
			\bottomrule			 
			\multirow{2}{*}{\textbf{M}}&\multicolumn{3}{c|}{\textbf{Different Views}} & \multirow{2}{*}{\textbf{BLCA}} & \multirow{2}{*}{\textbf{BRCA}} & \multirow{2}{*}{\textbf{LUAD}} & \multirow{2}{*}{\textbf{UCEC}}\\ \cline{2-4}
			&\textbf{\boldmath${P_s}$} & \textbf{\boldmath${P_m}$} & \textbf{\boldmath${P_l}$}& & & \\	
			
			\hline
			\multirow{7}{*}{\rotatebox{90}{\--{}}}          & \checkmark  &            &            & 0.631±0.035   & 0.646±0.037 & 0.644±0.008    & 0.701±0.021\\
			&             & \checkmark &            & 0.654±0.027   & 0.651±0.015 &  0.635±0.042   & \textbf{0.730±0.006}  \\
			&             &            &\checkmark  & 0.615±0.016   & 0.652±0.016 &  0.598±0.016   & 0.608±0.013 \\
			&\checkmark   & \checkmark &            & 0.637±0.021   & 0.656±0.007 &  0.661±0.007   & 0.694±0.011 \\
			&             & \checkmark & \checkmark & 0.663±0.003   & 0.648±0.032 &  0.657±0.009  & 0.691±0.006 \\
			& \checkmark  &            & \checkmark & 0.645±0.031   & 0.662±0.013 &  0.658±0.018  & 0.686±0.004 \\
			& \checkmark  & \checkmark &\checkmark  & 0.664±0.008   & 0.653±0.016 &  0.675±0.002   & 0.693±0.012 \\
			\hline
			\multirow{7}{*}{\rotatebox{90}{$R$}}  & \checkmark  &            &            & 0.681±0.012   & 0.633±0.028 &  0.645±0.011    & 0.687±0.036 \\
			&             & \checkmark &            & 0.654±0.009   & 0.637±0.013 &  0.655±0.007    & 0.679±0.022 \\			
			&             &            & \checkmark & 0.664±0.005   &0.616±0.016  &  0.647±0.013    & 0.663±0.036  \\		
			&\checkmark   & \checkmark &            & 0.674±0.002   &0.637±0.012  &  0.650±0.008    & 0.693±0.013 \\
			& \checkmark  &            &\checkmark  & 0.631±0.019   & 0.638±0.024 &  0.664±0.004    & 0.675±0.032 \\			
			&             & \checkmark & \checkmark & 0.683±0.003   & 0.628±0.041 &  0.646±0.031    & 0.661±0.046 \\
			& \checkmark  & \checkmark & \checkmark & 0.677±0.002   & 0.655±0.028 &  0.649±0.014    & 0.688±0.037 \\
			\hline
			
			\multirow{7}{*}{\rotatebox{90}{$C\&M$}}         & \checkmark  &            &            & 0.649±0.024   & 0.644±0.018  & 0.656±0.012    & 0.671±0.043 \\
			&             & \checkmark &            & 0.662±0.005   & 0.624±0.026  & 0.590±0.049    & 0.635±0.041 \\			
			&             &            & \checkmark & 0.671±0.003   & 0.628±0.035  &  0.580±0.042   & 0.608±0.031 \\
			& \checkmark  & \checkmark &            & 0.664±0.009   & 0.643±0.013  &  0.636±0.024   & 0.673±0.036 \\
			& \checkmark  &            & \checkmark & 0.651±0.017   & 0.605±0.035  &  0.661±0.008   & 0.646±0.046  \\
			&             & \checkmark & \checkmark & 0.664±0.003   & 0.634±0.027  &  0.568±0.046   & 0.630±0.039 \\
			& \checkmark  & \checkmark & \checkmark & 0.643±0.018   & 0.649±0.016  &  0.616±0.035   & 0.705±0.002 \\
			\hline
			
			\multirow{7}{*}{\rotatebox{90}{$R$\&$C\&M$}}         & \checkmark  &            &            & 0.661±0.007   & 0.655±0.005 &  0.660±0.011     & 0.689±0.025 \\			
			&             & \checkmark &            &0.671±0.002   & 0.621±0.034 & 0.629±0.032      & 0.680±0.014 \\			
			&             &            &\checkmark  &0.635±0.016    & 0.630±0.027  &  0.621±0.029    & 0.662±0.043  \\
			& \checkmark  & \checkmark &            & 0.655±0.012   & 0.643±0.022 & 0.665±0.016    & 0.690±0.008 \\
			& \checkmark  &            & \checkmark & 0.664±0.007   & 0.646±0.016 &  \textbf{0.678±0.005}    & 0.678±0.026 \\
			&             & \checkmark & \checkmark &0.649±0.012    & 0.636±0.025 & 0.643±0.017    & 0.650±0.046 \\
			& \checkmark  & \checkmark & \checkmark & \textbf{0.686±0.003}   & \textbf{0.679±0.005} & 0.672±0.008    & \textbf{0.730±0.002} \\
			\hline
		\end{tabular}}
		\caption{Performance of C-Index (mean) on four cancer datasets.}
		\label{tab:com}		
\end{table}

%\textbf{Impact of missing modalities.}To assess the significance of modal polymorphic data in AdaptS for survival prediction, we conducted a series of ablation experiments to investigate the effects of missing modal data. Illustrated in Fig.~\ref{fig:modality}, when solely utilizing data, the performance in predicting patient survival on the four cancer datasets is subpar. This might be due to the relatively small data volume, impacting the mapping relationship with survival-related data. AdaptS's performance in predicting using data alone is approximately 2\% to 3\% higher than when using data alone. In the and modes, AdaptS exhibits superior performance. Particularly noteworthy is that AdaptS achieved optimal performance across the four cancer cohorts when using multimodal data. However, while predicting patient survival is possible in the absence of multimodal data, maximizing the use of multimodal data significantly enhances the accuracy of survival prediction. This aligns with the broader trend in future research.

%\begin{figure}[!t]%[H]
	%\centering
	%\includegraphics[width=0.7\linewidth]{Figure3.pdf}
%	\centerline{\includegraphics[width=0.55\columnwidth]{Figure3.pdf}}
%	\caption{In the case of different modal combinations, AdaptS evaluates the survival prediction performance (C-index) through ablation experiments on four cancer datasets.}
%	\label{fig:modality}
	%\Description{Pipelines of robust watermarking, fragile watermarking, and our SepMark.}
%\end{figure}

\begin{figure}[!t]%[H]
	%\centering 
	%\includegraphics[width=0.7\linewidth]{Figure4.pdf}
	\centerline{\includegraphics[width=0.62\columnwidth]{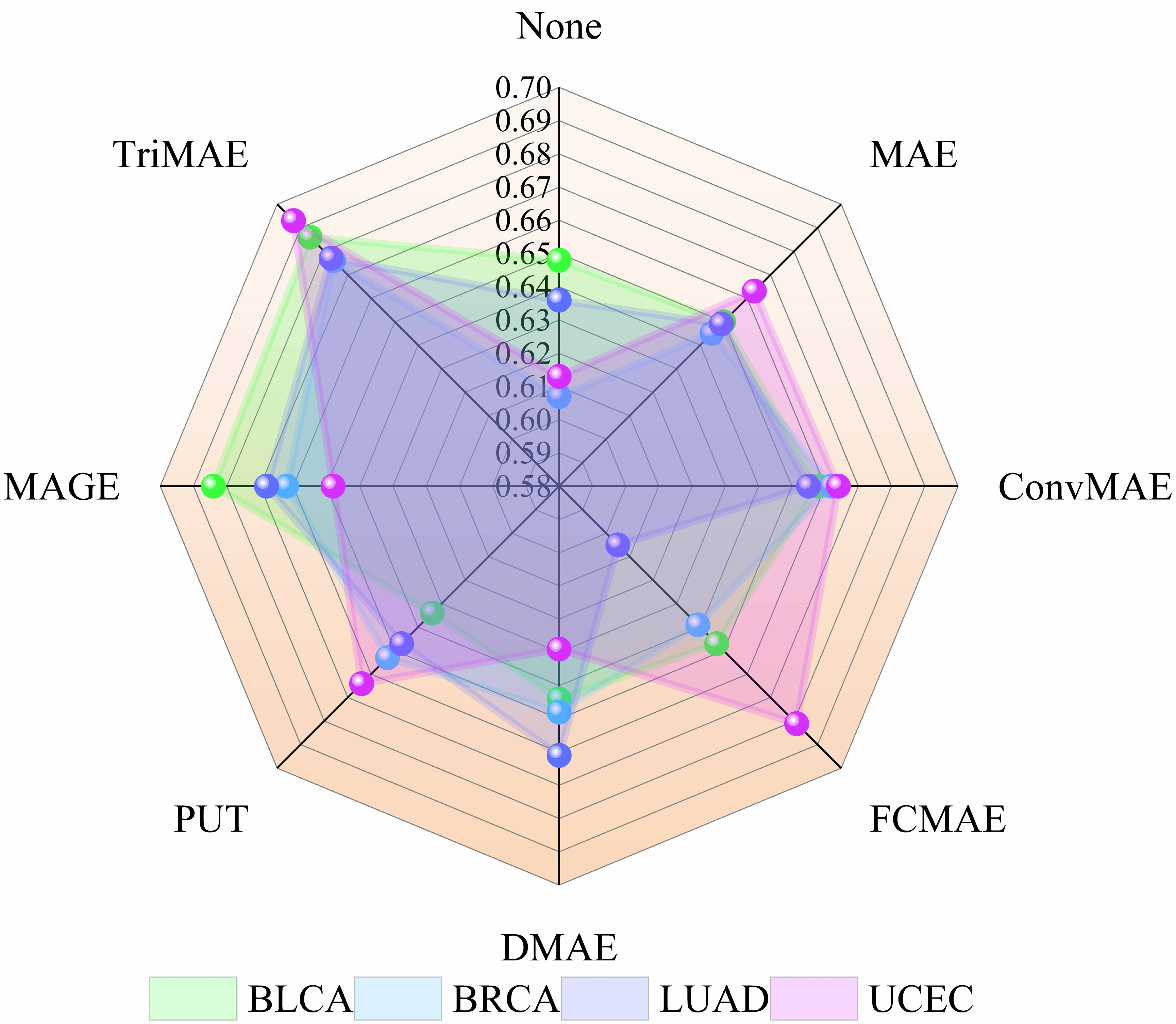}}
	\caption{Ablation experiments to evaluate the C-index performance of TriMAE on four cancer datasets in the absence of information within the modality.} 
	\label{fig:mae}
	%\Description{Pipelines of robust watermarking, fragile watermarking, and our SepMark.}
\end{figure}

\textbf{Effect of missing information within modalities.} To assess TriMAE's effectiveness, we conducted ablation experiments to investigate the impact of various MAEs on FORESEE's prediction performance.  Comparison methods included simple \textbf{MAE} \cite{he2022masked}, \textbf{ConvMAE} \cite{gao2022convmae}, \textbf{FCMAE} \cite{woo2023convnext}, \textbf{DMAE} \cite{wu2022denoising}, \textbf{PUT} \cite{liu2022reduce}, and \textbf{MAGE} \cite{li2023mage} strategies. Figure~\ref{fig:mae} illustrates that FORESEE without the use of a MAE is adversely impacted by missing data within the modality, resulting in poor prediction performance. Importantly, TriMAE achieves the best survival prediction results in comparative experiments. This may be attributed to the asymmetric mask design in TriMAE, which allows the model to leverage unmasked portions and masked tokens on different branches to obtain latent feature representations, addressing the issue of information loss in multimodal data. In conclusion, TriMAE is indispensable for multimodal survival prediction.%ConvMAE, MAGE, simple MAE, PUT, and DMAE all demonstrated favorable prediction performance. FCMAE exhibits poor performance when information within the modality is missing. TriMAE, however, demonstrated successful testing on four cancer cohorts and yielded the best predictive performance. This could be attributed to the structural design of the asymmetric mask in TriMAE, allowing it to leverage unmasked parts and masked tokens to derive potential feature representations on distinct branches, compensating for the information deficit in multimodal data. In summary, mask encoders are crucial for multimodal survival prediction. 

\begin{figure}[!t]%[H]
	%\centering
	%\includegraphics[width=0.8\linewidth]{Figure5.pdf}
	\centerline{\includegraphics[width=0.67\columnwidth]{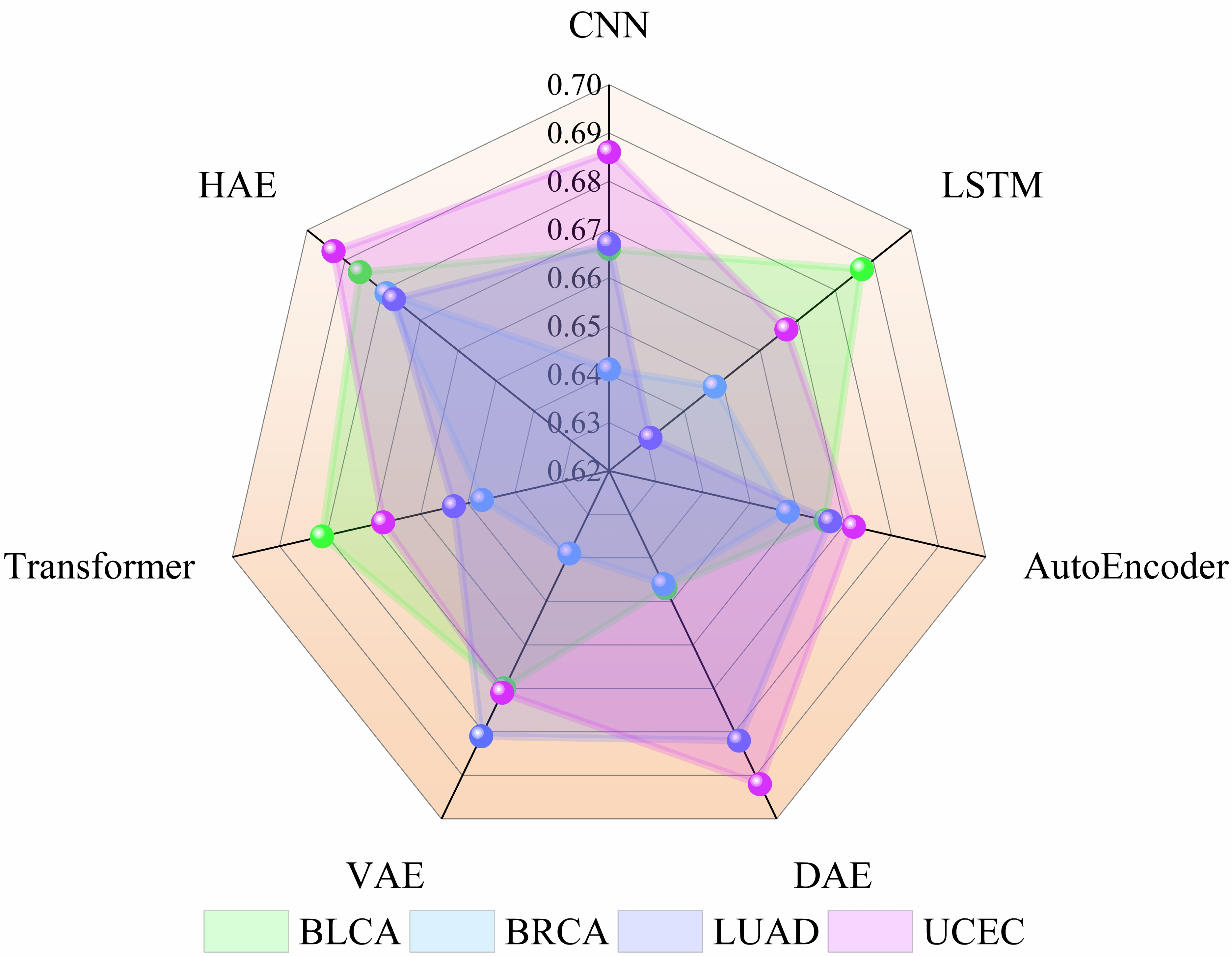}}
	\caption{Ablation experiments evaluating the C-index performance of the HAE on four cancer datasets.}
	\label{fig:hae}
	%\Description{Pipelines of robust watermarking, fragile watermarking, and our SepMark.}
\end{figure}

\textbf{Effect of HAE.} We compare HAE with several advanced molecular feature extraction methods, such as \textbf{CNN} \cite{gu2018recent}, \textbf{LSTM}, \textbf{AutoEncoder} \cite{vahdat2020nvae}, \textbf{DAE} \cite{du2016stacked}, \textbf{VAE} \cite{liang2018variational}, and \textbf{Transformer} \cite{beltagy2020longformer} strategies. Figure~\ref{fig:hae} illustrates the comparative survival prediction results of FORESEE across four cancer datasets.  Importantly, HAE achieved optimal results in the BLCA, BRCA, and LUAD datasets. This is attributed to HAE's CTA module, capable of capturing long-range dependencies among molecular features and acquiring local feature information. HAE's CNA module introduces global information from molecular data.  In summary, HAE exhibits significant advantages in molecular data extraction.

\section{Conclusion}
This paper proposes a novel end-to-end framework, FORESEE, for robustly predicting patient survival by mining multimodal information. The CFT effectively utilizes features at the cellular level, tissue level, and tumor heterogeneity level, correlating prognosis through a cross-scale feature cross-fusion method. The CTA and CNA modules in HAE can extract features from molecular data at both global and local perspectives. To address missing information within modalities, TriMAE can effectively reconstruct latent feature representations. Extensive experiments on four cancer datasets demonstrate that FORESEE significantly enhances the performance of survival prediction. To assess the impact of different fields of view, TriMAE, and HAE on the prediction performance of FORESEE, ablation experiments show that FORESEE achieves superior prediction performance, validating the necessity of our proposed modules.
%To address the specificity of diverse modalities data, this paper introduces a customized method for extracting information from multimodal data to predict cancer patient survival. Feature fusion across large, medium, and small fields of view facilitates the extraction of cross-scale contextual relationships by the model. The CTA and CNA module within HAE can concurrently extract features of molecular data from both global and local sources. To address the impact of missing information within modalities, TriMAE can effectively reconstruct latent feature representations. Extensive experiments conducted on four datasets of cancer patients have demonstrated that FORESEE significantly enhances the performance of survival prediction. To assess the impact of field of view, TriMAE, and HAE on the prediction performance of FORESEE, ablation experiments demonstrate that FORESEE attains superior prediction performance, validating the necessity of our proposed module.% This model holds significant potential for future applications in hospital settings.

%% The file named.bst is a bibliography style file for BibTeX 0.99c

\bibliographystyle{named}
\bibliography{ijcai23}
\end{CJK}
\end{document}